\documentclass[english]{cccconf}
\usepackage[comma,numbers,square,sort&compress]{natbib}
\usepackage{epstopdf}
\usepackage[T1]{fontenc}
\usepackage{amssymb}
\usepackage{amsmath}
\usepackage[linesnumbered,ruled,vlined]{algorithm2e}
\usepackage{bm}
\usepackage{graphicx}
\usepackage{subfigure}
\usepackage{multirow}
\usepackage{diagbox}
\usepackage{float}
\usepackage{color}
\begin{document}

\title{Global Update Guided Federated Learning}

\author{Qilong Wu\aref{at,eng,key},
	 Lin Liu\aref{dianji},
        Shibei Xue\aref{at,eng,key}\email{shbxue@sjtu.edu.cn}}



\affiliation[at]{Department of Automation, Shanghai Jiao Tong University, Shanghai 200240, P.~R.~China}
\affiliation[eng]{Key Laboratory of System Control and Information Processing, Ministry of Education of China, Shanghai 200240, P.~R.~China
}
\affiliation[key]{Shanghai Engineering Research Center of Intelligent Control and Management, Shanghai 200240, P.~R.~China
        }
\affiliation[dianji]{ School of Electronic Information Engineering, Shanghai Dianji University, Shanghai 201306, P.~R.~China
}
\maketitle

\begin{abstract}
Federated learning protects data privacy and security by exchanging models instead of data. However, unbalanced data distributions among participating clients compromise the accuracy and convergence speed of federated learning algorithms. To alleviate this problem, unlike previous studies that limit the distance of updates for local models, we propose global-update-guided federated learning (FedGG), which introduces a model-cosine loss into local objective functions, so that local models can fit local data distributions under the guidance of update directions of global models. Furthermore, considering that the update direction of a global model is informative in the early stage of training, we propose adaptive loss weights based on the update distances of local models. Numerical simulations show that, compared with other advanced algorithms, FedGG has a significant improvement on model convergence accuracies and speeds. Additionally, compared with traditional fixed loss weights, adaptive loss weights enable our algorithm to be more stable and easier to implement in practice.
\end{abstract}



\footnotetext{This work was supported in part by the National Natural Science Foundation of China (NSFC) under Grants 61873162, 61973317, and in part by the Open Research Project of the State Key Laboratory of Industrial Control Technology, Zhejiang University, China (No. ICT2022B47), and in part by the Scientific Research Funding of Shanghai Dianji University (No. B1-0288-21007-01-023).}

\section{Introduction}
With the rapid development of edge devices, e.g., mobile phones and wearable devices, unprecedented amounts of data are generated. With these data, machine learning has made breakthrough progresses and provided many successful applications, e.g., next word prediction \cite{Yang}, cardiac health monitoring \cite{Kumar}, load forecasting \cite{Guan,Jia}, pose estimation \cite{GuanQ,GuanQ1} and autonomous driving \cite{Sheng,Sheng1,Sheng2}, etc. In these applications, traditional methods for machine learning are centralized where corresponding data is  aggregated and calculated on a server \cite{AbdulRahman}. However, due to privacy concerns and data protection regulations \cite{Voigt}, it becomes difficult to implement these centralized methods in practice.

In order to solve the above problems, federated learning (FL) \cite{McMahan} emerges as a new paradigm, which constructs a shared model on a server by aggregating local models learned from training data of each client. Since model training is performed locally on clients and no data is transmitted, FL can comprehensively protect privacy of clients and fully utilize computational resources of those edge devices. However, compared to distributed learning in data center settings and traditional private data analysis, FL is challenged by statistical heterogeneity, where the data on clients is not identically or independently distributed (Non-IID) \cite{Li}.
Ref. \cite{Lix} points out that facing Non-IID data, the accuracy and the convergence rate of Federated Averaging (FedAvg) \cite{McMahan}, the most basic method in federated learning, degrades significantly. In FedAvg with Non-IID data, although the number of communication rounds is reduced by increasing the number of iterations for the training on clients, a large number of epochs may cause each client to achieve its local optimum rather than a global optimum such that the algorithm diverges.
Therefore, many studies devote to enhancing the performance of FedAvg on Non-IID data by improving local training.

For improving local trainings, FedProx \cite{Li1} introduces a proximal term to penalize local objective functions when a $l_{2}$ distance between local and global models becomes large. However, in FedProx, penalty coefficients are the same for different parameters in the proximal term, and the importance of the parameters often varies in different tasks.
Inspired by lifelong learning~\cite{Hou}, FedCurv \cite{Shoham} employs a penalty to the parameters that deviate from other clients with a large amount of information. However, it is necessary to transmit Fisher information of clients to server, which increases communication costs. To tackle this problem, FedCL \cite{Yao} introduces a small auxiliary dataset on the server with which the Fisher information of the global model is calculated to obtain the penalty coefficient matrix. However, the dataset on the server side leaks privacy. Instead of adding constraints to a local objective, Ref.~\cite{Karimireddy} introduces global and local control variables to revise local updates, which doubles communication costs. However, as shown in Ref.~\cite{Li3}, these approaches fail to achieve a good performance on image datasets, which can be as bad as FedAvg.

%

In this paper, instead of limiting the distance of updates for local models in most methods, we propose global-update-guided federated learning (FedGG), which guides local updates by maximizing a cosine similarity between global and local updates.  
In this way, we can reduce model  divergence between clients and improve the generalization ability of federated models. Unlike FedCurv, FedCl and SCAFFOLD methods, our method does not introduce additional communication costs.
Furthermore, considering that the update direction of the global model contains different amounts of information in different periods of training, we propose adaptive weights based on the distances of updates for local models, which makes our algorithm more flexible and stable.
Our numerical experiments show that, compared with other advanced algorithms, FedGG has an accuracy improvement of $0.5\%$, $2.2\%$ and $2.9\%$ on Non-IID SVHN, CIFAR-10 and CIFAR-100 datasets. Meanwhile, the speedup in communication efficiency can achieve $2.63$ times at most.

\section{Preliminaries}

\subsection{Problem Statement of  Federated Learning}
 Federated learning aims to learn a consensus model in a decentralized manner. Concretely speaking, considering $N$ clients $C_{1},...,C_{N}$ where each client $C_{i}$ possesses a local dataset $\mathcal{D}_{i}$, federated learning is to learn a global model $w$ over the dataset $\mathcal{D}\triangleq\cup_{i\in[N]}\mathcal{D}_{i}$ with the coordination of a central server, while raw data remains in clients. An optimization description for federated learning is
 \begin{equation}
 	\label{eq1}
 	\mathop{\arg\min}\limits_{w}\ \ \mathcal{L}(w)=\sum_{i=1}^N\dfrac{|\mathcal{D}_{i}|}{|\mathcal{D}|}L_{i}(w),
 \end{equation}
where $|\cdot|$ denotes the cardinality of sets and $L_{i}(w) = \mathbb{E}_{(x,y)\sim\mathcal{D}_{i}}[\ell_{i}(w;(x,y))]$ is the expectation for the cross-entropy loss function $\ell_{i}$ of the client $C_{i}$ in classification problems. Here, $(x,y)$ denotes the input data $x$ and its corresponding label $y$~\cite{McMahan}.
\subsection{Federated Averaging Algorithm}
As the first algorithm in federated learning, federated Averaging (FedAvg) algorithm requires all clients to share the same training configurations from optimizers to learning rates. In each communication round $r$, a subset of clients $S_{r}$ with $|S_{r}|\leq N$ is selected and the central server sends a global model $w^{r}$ to them. Each selected client locally executes the stochastic gradient descent (SGD) algorithm for $\tau$ times, where $w^{r}$ is taken as an initial model. This local model is updated as
 \begin{equation}
 		\label{eq2}
w_{i}^{r}(m)=w_{i}^{r}(m-1)-\eta\nabla L_{i}(w_{i}^{r}(m-1))
\end{equation}
by optimizing a local objective $L_{i}(w)$,
where $m=1,...,\tau$, $w_{i}^{r}(0)=w^{r}$, $\eta$ is the learning rate, and $\nabla L_{i}(\cdot)$ is the gradient at the client $C_{i}$. Since the batch stochastic gradient descent algorithm is usually used in practice, the number of updates $\tau$ can be computed as $\tau=|\mathcal{D}_{i}|E/B$, where $B$ is the mini-batch size and $E$ is the number of epochs. In the above FedAvg algorithm, we require $\tau>1$. Note that when $\tau=1$, this algorithm reduces to FedSGD \cite{McMahan}.

After $\tau$ times local updates, each client sents $w_{i}^{r}(\tau)$ to the server which updates the global model $w^{r+1}$ as
 \begin{equation}
 		\label{eq3}
w^{r+1}=\sum_{i\in S_{r} }\frac{|\mathcal{D}_{i}|}{|\mathcal{D}_{S_{r}}|} w_{i}^{r}(\tau),
\end{equation}
with $\mathcal{D}_{S_{r}}\triangleq\cup_{i\in[S_{r}]}\mathcal{D}_{i}$.

\section{Global-Update-Guided Federated
Learning Algorithm}
Unlike previous methods for improving FedAvg by limiting the local distance for updates, we address data heterogeneity by correcting the direction of updates for local models. Furthermore, given that the update information of a global model is of importance in different stages of training, we introduce an adaptive loss weight mechanism based on the distance of updates for the local model to improve the adjustable range of algorithm parameters and the stability of our algorithm. With the above considerations, we propose global-update-guided federated learning (FedGG).
\subsection{Measurement of the Similarity of Updates}
The main idea of our method is to guide the update of local models along the direction of updates for a global model. That is, we want the local update direction to be highly similar to the global one. In this paper, we introduce a cosine function to evaluate the similarity between variations of two models, which is often used in text classifications \cite{LiB}.

First we need to determine the representation of the direction of updates for the global model and local models, respectively.
To represent it of the global model, we use the differences between the last two global models received by clients, i.e.,
\begin{equation}
	\begin{aligned}
		\label{eq99}
\triangle w^{r-1}=w^{r}-w^{r-1}
	\end{aligned}	
\end{equation}
at communication round $r$ instead of an intuitive idea of using global gradients. This is because the global gradients is difficult to obtain and we are also required to protect data privacy. To show this, we can rewrite Eq.~(\ref{eq3}) as
 \begin{equation}
\begin{aligned}
	\label{eq4}
	&w^{r+1}=w^{r}+\triangle w^{r} \\
		&\triangle w^{r}=\sum_{i\in S_{r} }\frac{|\mathcal{D}_{i}|}{|\mathcal{D}_{S_{r}}|} \left[w_{i}^{r}(\tau)-w^{r}\right].
\end{aligned}	
\end{equation}
Only when $\tau=1$, the update process of the server can be considered as a stochastic gradient descent with a learning rate 1 and $\triangle w^{r}$ represents the global gradient. This means that the client needs to upload the model after each iteration, which defeats the original purpose of FedAvg for reducing communication costs through multiple local iterations. Nevertheless, when the number of local iterations $\tau>1$, we can use $\triangle w^{r}$ to approximate the global gradient. However, in the communication round $r$, the participating clients cannot get $\triangle w^{r}$, because $\triangle w^{r}$ can only be obtained after performing local updates. Hence, given the above issues, $\triangle w^{r-1}$ can be a good choice to represent the global model update direction, since it's similar to $\triangle w^{r}$ especially in the early stage of communications, and can be easily obtained from $\triangle w^{r-1}=w^{r}-w^{r-1}$ for clients participating in federated learning.

For local models, we use the differences between the current local models and the global model to represent its direction for updates, i.e.,
 \begin{equation}
	\begin{aligned}
		\label{eq100}
\triangle w_{i}^{r}(m-1)=w_{i}^{r}(m-1)-w^{r}
	\end{aligned}
\end{equation}
for the client $C_{i}$ at the $m_{th}$ local iteration, but not the gradient of the local model. This is because if the gradient of the local model is directly introduced into the objective function, the second derivative for the update in the stochastic gradient descent algorithm would greatly increase computational costs.

Hence, with the expressions of the global and local model updates, for the client $C_{i}$ at the $m_{th}$ local iteration, we can evaluate the similarity of the changed direction  between the global model and the local model by the cosine similarity
 \begin{equation}
 	\begin{aligned}
	\label{eq5}
	cos_{i}^{r}(\theta_{m-1})&=
	\dfrac{\left \langle \triangle w^{r-1} , \triangle w_{i}^{r}(m-1) \right \rangle}
	{\left \| \triangle w^{r-1} \right \|
	\left \| \triangle w_{i}^{r}(m-1) \right \|}
\\
&=
\dfrac{\left \langle w^{r}-w^{r-1} ,
	 w_{i}^{r}(m-1)-w^{r} \right \rangle}
{\left \|  w^{r}-w^{r-1} \right \|
	\left \|  w_{i}^{r}(m-1)-w^{r} \right \|},
\end{aligned}
\end{equation}
where $\left \langle \cdot,\cdot \right \rangle$ denotes the inner product between vectors, $\left \| \cdot \right \|$ denotes the $l_{2}$ distance of a vector, $\theta_{m-1}$ denotes the angle between the update vectors of the global and local models at the $m_{th}$ iteration. The value of the cosine similarity of the directions of updates for the global and local models lies in between 0 and 1. The greater the value is, the more similar the update directions of the global and local models are.
\subsection{Local Training Objective}
The basic idea of our method is to guide the update of the local model along the update direction of the global model. Thus, a local objective function can be divided into two parts. The first part is the original local loss function, which can be a classic supervised learning loss (e.g., cross-entropy loss) denoted as $\ell_{sup}$. The second part is used to guide the update of the local model, which is a model-cosine loss $\ell_{cos}$. Since the more similar the updates of the models, the smaller this loss value should be, in the communication round $r$, we define the model-cosine loss $\ell_{cos}$ for the client $C_{i}$ as
 \begin{equation}
	\begin{aligned}
		\label{eq6}
	\ell_{cos}
	&=1-cos_{i}^{r}(\theta)
	\\
	&=1-\dfrac{\left \langle w^{r}-w^{r-1} ,
		w_{i}^{r}-w^{r} \right \rangle}
	{\left \|  w^{r}-w^{r-1} \right \|
		\left \|  w_{i}^{r}-w^{r} \right \|}.
	\end{aligned}
\end{equation}
Thus, the total loss for an input $(x,y)$ is computed as
 \begin{equation}
	\begin{aligned}
		\label{eq7}
\ell=\ell_{sup}(w_{i}^{r};(x,y))+\lambda \ell_{cos}(w_{i}^{r};w^{r};w^{r-1}),
	\end{aligned}
\end{equation}
where $\lambda$ is a hyper-parameter to adjust the weights of the model-cosine loss. Hence, the local optimization problem is mathematically described as
 \begin{equation}
 		\label{eq8}
	\begin{aligned}
		 	\mathop{\min}\limits_{w_{i}^{r}}\ \  \mathbb{E}_{(x,y)\sim\mathcal{D}_{i}}\left[ \ell_{sup}(w_{i}^{r};(x,y))+\lambda \ell_{cos}(w_{i}^{r};w^{r};w^{r-1})\right].
	\end{aligned}
\end{equation}
With the local objective~(\ref{eq8}), we force the local model $w_{i}^{r}$ to fit the local data distribution with the help of the direction of the global model update, which alleviates the discrepancy between the local and global models under the Non-IID condition, and improves the generalization ability of the model.
\subsection{Adaptive Loss Weights}
To further explore the guiding role of model-cosine loss on local model update, we calculate the gradient of the local objective function $\ell$ with respect to the local model $w_{i}^{r}$ as
 \begin{equation}
	\begin{aligned}
		\label{eq9}
		\nabla\ell
		&=\nabla\ell_{sup}+\lambda\nabla \ell_{cos}
		\\
		&=\nabla\ell_{sup}+\lambda\nabla\left[1-\dfrac{\left \langle w^{r}-w^{r-1} ,
			w_{i}^{r}-w^{r} \right \rangle}
		{\left \|  w^{r}-w^{r-1} \right \|
			\left \|  w_{i}^{r}-w^{r} \right \|} \right]
		\\
		&=\nabla\ell_{sup}-\dfrac{\lambda}{\left \|  w_{i}^{r}-w^{r} \right \|}\left[\right.\dfrac{w^{r}-w^{r-1}}{ \left \|  w^{r}-w^{r-1} \right \| }
		\\
		&\ \ \ \  +cos_{i}^{r}(\theta)\dfrac{w_{i}^{r}-w^{r}}{\left \|  w_{i}^{r}-w^{r} \right \|} \left.\right]
	\end{aligned}
\end{equation}
where $cos_{i}^{r}(\theta)=\frac{\left \langle w^{r}-w^{r-1} ,
	w_{i}^{r}-w^{r} \right \rangle}
{\left \|  w^{r}-w^{r-1} \right \|
	\left \|  w_{i}^{r}-w^{r} \right \|}$,
and for brevity, we denote $\ell_{sup}(w_{i}^{r};(x,y))$ and $\ell_{cos}(w_{i}^{r};w^{r};w^{r-1})$ as $\ell_{sup}$ and $\ell_{cos}$, respectively.

Note that compared with FedAvg, FedGG introduces two vectors in the same direction as the normalized vectors $\frac{w^{r}-w^{r-1}}{ \left \|  w^{r}-w^{r-1} \right \| }$ and $\frac{w_{i}^{r}-w^{r}}{\left \|  w_{i}^{r}-w^{r} \right \|} $ to correct the local update of the client, and the lengths are $\frac{\lambda}{\left \|  w_{i}^{r}-w^{r} \right \|}$ and $\frac{\lambda cos_{i}^{r}(\theta)}{\left \|  w_{i}^{r}-w^{r} \right \|}$ respectively. The former is the update direction for the global model, which is used to guide the update of the local model. The latter is the direction in which the current local model is updated from the global model. Also, its length has an extra coefficient $cos_{i}^{r}(\theta)$ compared to the former, which is to evaluate the update direction. This is because local training also needs to fit the local data distribution such that the local update direction would not be the same as the global model update direction. If the direction is similar to the direction of the global update, a larger weight is applied, and vice versa.

Furthermore, we note that if the weight-tuning parameter $\lambda$ is set to be a constant, the coefficient $\frac{\lambda}{\left \|  w_{i}^{r}-w^{r} \right \|}$ decreases as the distance $\left \|  w_{i}^{r}-w^{r} \right \|$ increases. However, according to the results in Ref.~\cite{Geyer}, with the increase of communication rounds, the update scale of local models gradually decreases. This means in this case of a constant $\lambda$ the coefficient is larger in the early stage of training and smaller in the later stage of training. This is unexpected because a large number of experiments \cite{Li3} show that the update direction of the global model is more informative in the early stage. And the varied direction of the global model is less helpful in the later stage of training since the global model is close to convergence. Hence, we consider an adaptive design for the hyper-parameter $\lambda$, which is defined as
 \begin{equation}
	\begin{aligned}
		\label{eq10}
		\lambda=\mu
        \left \|  w_{i}^{r}(m-1)-w^{r} \right \|
		\left \|  w_{i}^{r}(m-1)-w^{r}(m-2) \right \|
	\end{aligned}
\end{equation}
at the $m_{th}$ local iteration, which can adjust the weights of the model-cosine loss. Note that we require $m > 2$ since at the first local iteration, the current local model is initialized by the global model so that there is no updates and no model-cosine loss.

Although $\lambda$ relates to the local model at the current moment, in order to reduce the local computational cost, we treat it as a constant, i.e.,$\nabla\lambda=0$. Thus, at the $m_{th}$ local iteration, we take the adaptive loss weights Eq.~(\ref{eq10}) into Eq.~(\ref{eq9}) and thus obtain
 \begin{equation}
	\begin{aligned}
		\label{eq11}
		\nabla\ell
		&=\nabla\ell_{sup}-\mu\left \|  w_{i}^{r}(m-1)-w^{r}(m-2) \right \|
		\times \\&
		\left[\right.\dfrac{w^{r}-w^{r-1}}{ \left \|  w^{r}-w^{r-1} \right \| }
	 +cos_{i}^{r}(\theta_{m-1})\dfrac{w_{i}^{r}(m-1)-w^{r}}{\left \|  w_{i}^{r}(m-1)-w^{r} \right \|} \left.\right].
	\end{aligned}
\end{equation}
So far, by introducing adaptive weights, the weights of the two directions that play a guiding role in the local update will be determined by the distance of updates for the latest local iteration, which is expected that the weight of model-cosine loss decreases as the number of communication rounds increases.

In summary, our FedGG algorithm is described in Algorithm \ref{Algorithm 1} without sampling mechanism. FedGG is still applicable in federated learning with the participation of non-persistent clients. In the case, we simply replace $w^{r-1}$ with the latest received global model except this round when calculating the direction of the global model update. Note that on line 19 in Algorithm \ref{Algorithm 1}, we adopt $w_{i}^{r}(m)\leftarrow w_{i}^{r}(m-1)-\eta(\nabla\ell_{sup}+\lambda\nabla \ell_{cos}) $ for the local update instead of $w_{i}^{r}(m)\leftarrow w_{i}^{r}(m-1)-\eta\nabla\ell$, because we treat $\lambda$ as a constant to reduce the local computational cost.
\IncMargin{1em}
\begin{algorithm} \SetKwData{Left}{left}\SetKwData{This}{this}\SetKwData{Up}{up} \SetKwFunction{Union}{Union}\SetKwFunction{FindCompress}{FindCompress} \SetKwInOut{Input}{Input}\SetKwInOut{Output}{Output}
	\Input{number of communication rounds $R$, number of clients $N$, number of local
		epochs $E$, learning rate $\eta$, hyper-parameter $\mu$}
	\Output{the final model $w^{R}$}
	\BlankLine
	\textbf{Server executes:}\\
	initialize $w^{0}$\\
	\For{$r\leftarrow 0$ \KwTo $R-1$} {
		\For{$i\leftarrow 1$ \KwTo $N$ in parallel}{send the global model $w^{r}$ to client $C_{i}$\\
		waitting for $w_{i}^{r}\leftarrow$ \textbf{ClientLocalTraining}(i, $w^{r}$)}
	server updates: $w^{r+1}=\sum_{i=1 }^{N}\frac{|\mathcal{D}_{i}|}{|\mathcal{D}|} w_{i}^{r}$\\
	}
	return $w^{R}$\\
	\textbf{ClientLocalTraining}(i, $w^{r}$):\\
	$w_{i}^{r}(0)\leftarrow w^{r}$\\
	\For{$m \leftarrow 1$ \KwTo
		 $\tau=\frac{|\mathcal{D}_{i}|}{B}E$}{
		 $\lambda\leftarrow0$ \\ $\ell_{cos}\leftarrow0$\\
		$\ell_{sup}\leftarrow$ \textit{CrossEntropyLoss}($F_{w_{i}^{r}(m)}(x)$, $y$)
		\\	
		\If{communication round $r\neq0$ \textbf{and} $m>1$ }{
			$\ell_{cos}\leftarrow1-\frac{\left \langle w^{r}-w^{r-1} ,
				w_{i}^{r}(m-1)-w^{r} \right \rangle}
			{\left \|  w^{r}-w^{r-1} \right \|
				\left \|  w_{i}^{r}(m-1)-w^{r} \right \|}$
			\\
			$\lambda\leftarrow\mu\left \|  w_{i}^{r}(m-1)-w^{r} \right \|
			\left \|  w_{i}^{r}(m-1)-w^{r}(m-2) \right \|$}	
	$	\ell\leftarrow\ell_{sup}+\lambda\ell_{cos}$\\
	$w_{i}^{r}(m)\leftarrow w_{i}^{r}(m-1)-\eta(\nabla\ell_{sup}+\lambda\nabla \ell_{cos}) $
	}
$w_{i}^{r}\leftarrow w_{i}^{r}(m)$\\
return  $w_{i}^{r}$ to server

	\caption{Global-Update-Guided Federated Learning, FedGG}
    \label{Algorithm 1}	
\end{algorithm}
\DecMargin{1em}

\section{Experiments}
\subsection{Experimental Setup}
We compare FedGG with three state-of-the-art federated learning algorithms including FedAvg, FedProx, and Scaffold. In order to evaluate the performance of our algorithm, we test it on three different datasets, namely SVHN (99289 images with 10 classes), CIFAR-10 (60,000 images with 10 classes) and CIFAR-100 (60,000 images with 100 classes). Due to the difficulty in classification for these data sets ,  training networks with different complexities are selected. For SVHN, we adopt a three-layer MLP with ReLU activation, where the dimensions of the input layer, hidden layer and output layer are 3$\times$32$\times$32, 512, and 10, respectively. For CIFAR-10, we use a simple CNN network, which is composed of two 5$\times$5 convolution layers with 6 and 16 channels, respectively. They are followed by a 2$\times$2 maxpooling and two fully connected layers whose dimensions are 120 and 84 , respectively. This network takes ReLU as an activation function. For CIFAR-100, we use VGG-16 for feature extraction and classification. For fair comparison, we adopt the same network structure for all algorithms.

\begin{figure}[htb]
	\centering
	\subfigure[SVHN]{
		\begin{minipage}[t]{0.33\linewidth}
			\centering
			\includegraphics[width=\hsize]{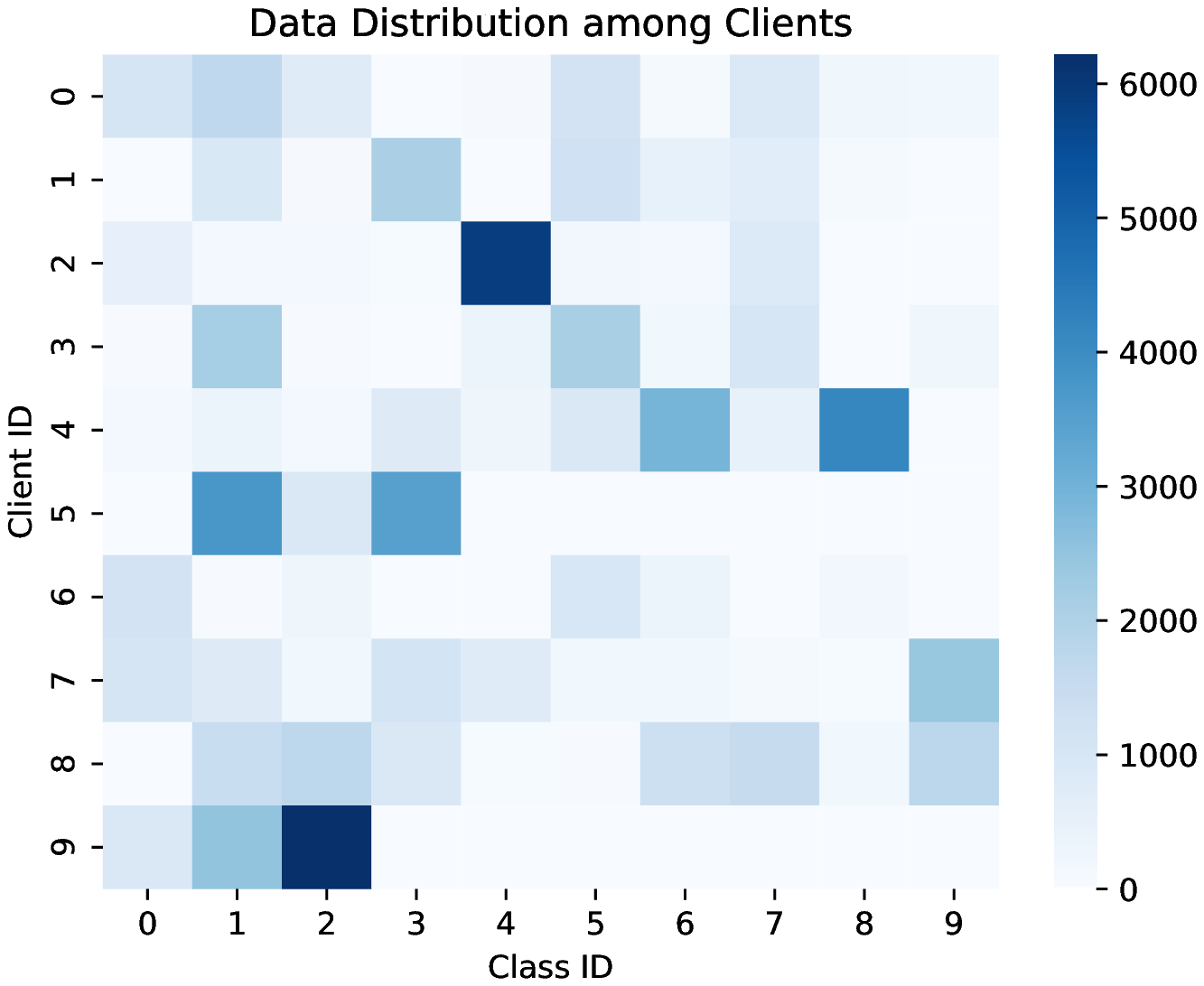}
		\end{minipage}%
	}%
	\subfigure[CIFAR-10]{
		\begin{minipage}[t]{0.33\linewidth}
			\centering
			\includegraphics[width=\hsize]{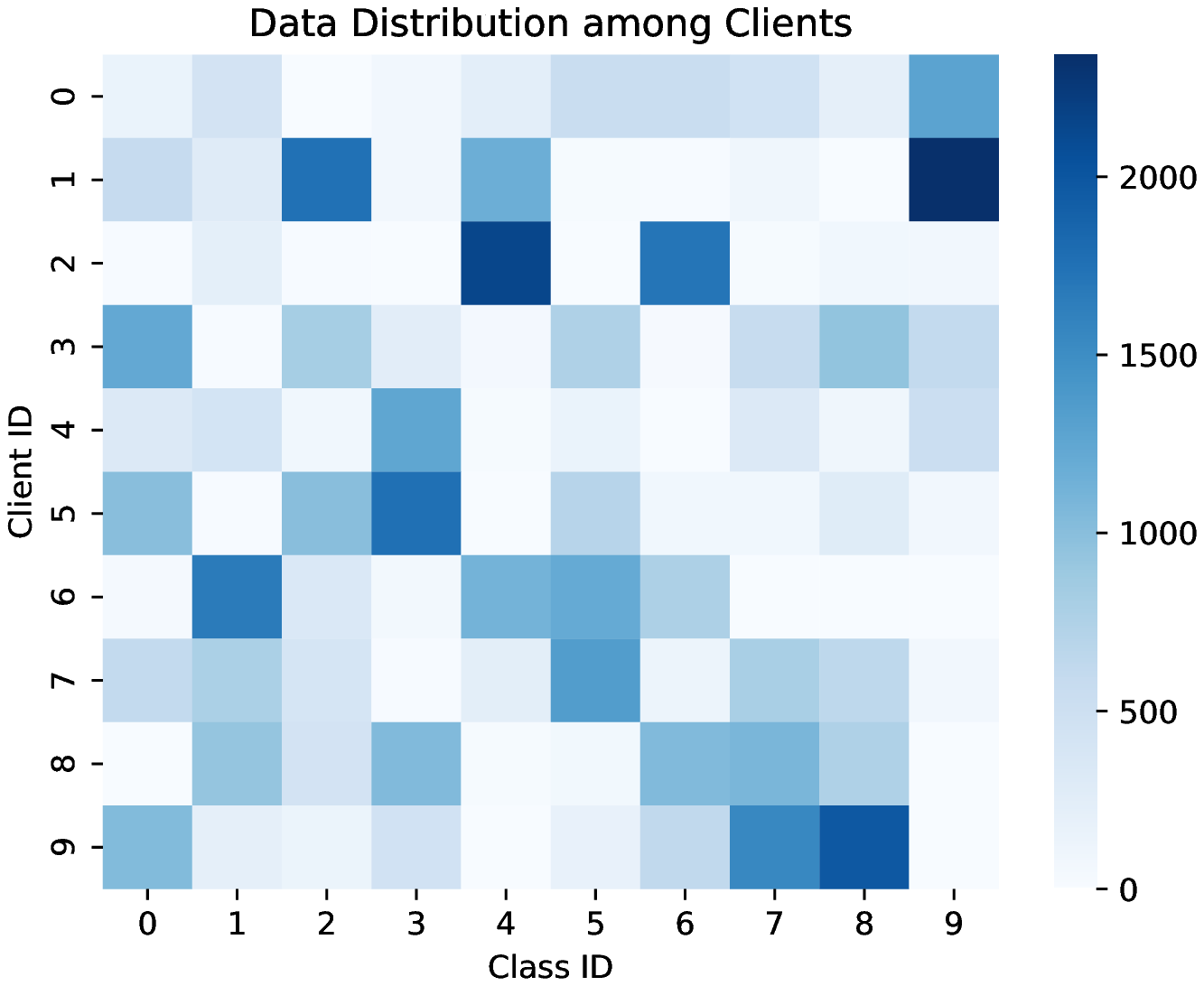}
		\end{minipage}%
	}%
	\subfigure[CIFAR-100]{
		\begin{minipage}[t]{0.33\linewidth}
			\centering
			\includegraphics[width=\hsize]{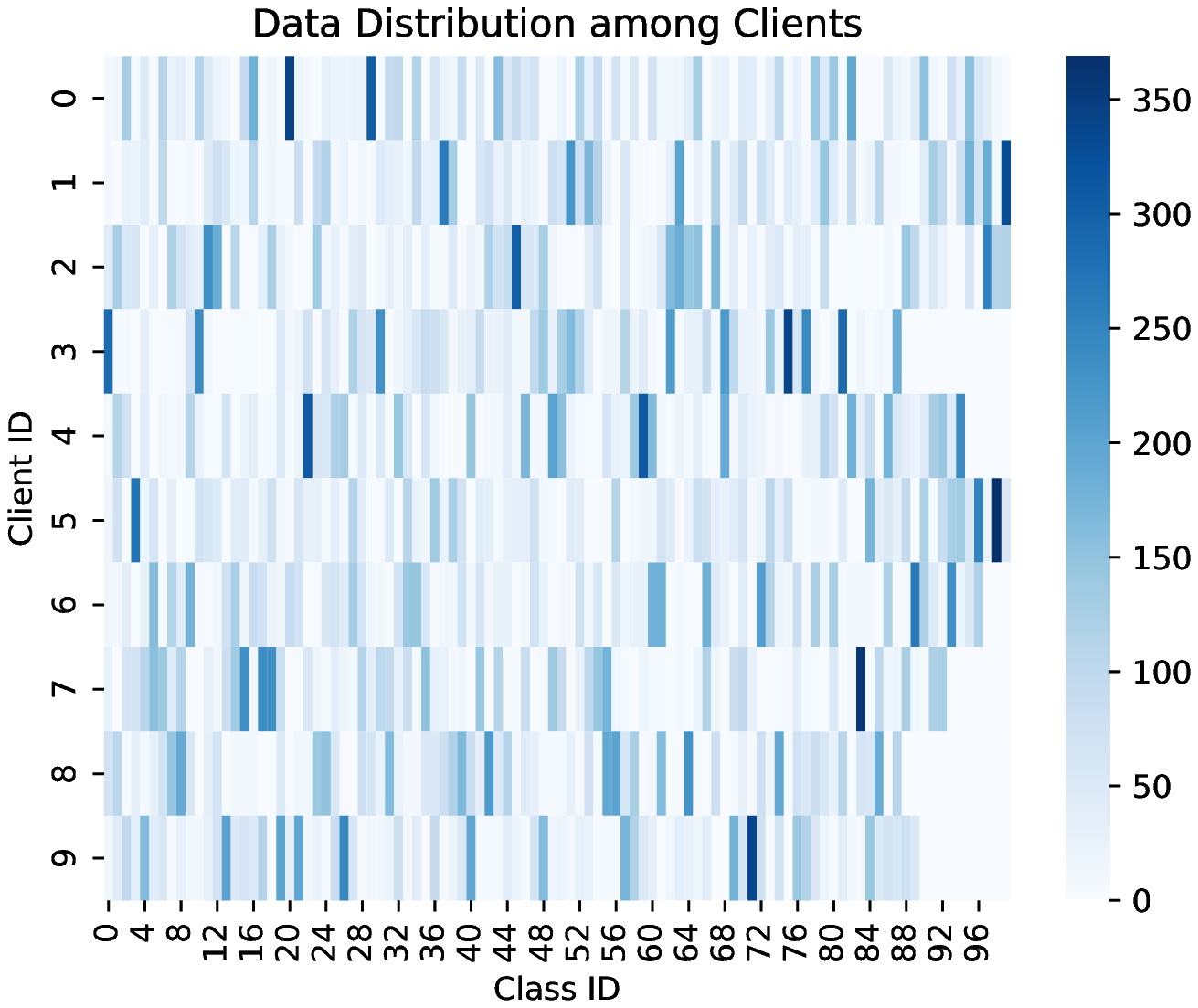}
		\end{minipage}
	}%
	\centering
	\caption{Distribution of three data sets on 10 clients with $\beta=0.5$.}
	\label{fig1}
\end{figure}

In order to simulate the Non-IID  data in the real world, similar to previous studies in Refs. \cite{Yao,Li3}, we use vector $\bm{p_{i}}$ with $p_{ij}\geq0$, $j\in[1,M]$ and $\Vert \bm{p_{i}} \Vert_1=1$ to represent the distribution of $M$ types of data for each client, where $\bm{p_{i}}\sim Dirichlet(\beta)$ and the parameter $\beta>0$ can adjust the degree of Non-IID data. A less $\beta$ indicates a more unbalanced distribution of the data. Taking $\beta=0.5$ as an example, we show the distribution of data on 10 clients in Fig.~\ref{fig1}.
We use Pytorch to implement FedGG and other algorithms. For training models, we use the SGD optimizer with a learning rate 0.01 and an SGD momentum 0.9. We set the number of epochs $E=5$ and the number of clients $N=10$ in each communication round. Besides, the batch size $B=64$ and the number of communication round $R$ is set to 100, at which point all algorithms barely improved on the three datasets.
\subsection{Comparison on Accuracy }
We evaluate the performance of the four algorithms on the three datasets with $\beta=0.5$ and $\beta=0.1$ and each algorithm is repeated by three times. For FedGG, we let the hyperparameters $\mu=\left\lbrace 001,0.01,0.1,1\right\rbrace$ and show that it achieves the best performance in all three data sets with $\mu=0.01$. For FedProx, we set its hyperparameter to be 0.01, which is the best hyperparameter mentioned in Ref.~\cite{Li1}. The highest tset accuracies of these algorithms over 100 communication rounds are given in Table~\ref{tab1} and Table~\ref{tab2}.
\begin{table}[htb]
	\centering
	\caption{Comparison on the highest test accuracy for the four algorithm in the case of $\beta=0.5$ }
	\label{tab1}
	\resizebox{\linewidth}{!}{
	\begin{tabular}{c|c|c|c}
		\hhline
		Method          &SVHN &CIFAR-10 & CIFAR-100  \\ \hline
		FedGG     & $\bm{80.0\%}\pm0.2\%$ & $\bm{71.1\%}\pm0.2\%$  & $\bm{61.6\%}\pm0.1\%$ \\ \hline
		FedAvg    & $79.6\%\pm0.3\%$ & $69.2\%\pm0.4\%$  & $60.1\%\pm0.2\%$ \\ \hline
		Fedprox   & $79.7\%\pm0.4\%$ & $69.0\%\pm0.4\%$  & $59.8\%\pm0.2\%$ \\ \hline
		Scaffold  & $78.2\%\pm0.3\%$ & $70.1\%\pm0.3\%$  & $61.0\%\pm0.1\%$ \\ \hline
	\end{tabular}}
\end{table}
\begin{table}[htb]
	\centering
	\caption{Comparison on the highest test accuracy for the four algorithm in the case of $\beta=0.1$ }
	\label{tab2}
\resizebox{\linewidth}{!}{
	\begin{tabular}{c|c|c|c}
		
		\hhline
		Method          &SVHN &CIFAR-10 & CIFAR-100  \\ \hline
		FedGG     & $\bm{70.8\%}\pm0.3\%$ & $\bm{63.8\%}\pm0.1\%$  & $\bm{55.2\%}\pm0.2\%$ \\ \hline
		FedAvg    & $70.3\%\pm0.2\%$ & $61.6\%\pm0.3\%$  & $52.0\%\pm0.3\%$ \\ \hline
		Fedprox   & $70.1\%\pm0.3\%$ & $60.5\%\pm0.2\%$  & $50.0\%\pm0.2\%$ \\ \hline
		Scaffold  & $64.4\%\pm0.2\%$ & $61.2\%\pm0.3\%$  & $53.3\%\pm0.2\%$ \\ \hline
	\end{tabular}}
\end{table}

From the results in Table~\ref{tab1} and Table~\ref{tab2}, it can be concluded that FedGG achieves the highest test accuracy in all three data sets for both $\beta=0.5$ and $\beta=0.1$. Especially, in the extremely unbalanced case $\beta=0.1$; i.e., the data distribution is closer to a practical situation, FedGG has a better performance, which is respectively $0.5\%$, $2.2\%$ and $2.9\%$ higher than the second highest test accuracy in the three data sets. Furthermore, we notice that the performance of the same algorithm on the same data set degrades in the case of $\beta=0.1$ compared to $\beta=0.5$, since the model divergence is more critical between clients. By introducing the model-cosine loss, the local model update is guided by the direction of global model update, which alleviates this problem.
\subsection{Comparison on Communication Efficiency}
\begin{figure*}
	\centering
	\subfigure[SVHN]{
		\begin{minipage}[h]{0.3\linewidth}
			\centering
			\includegraphics[width=\hsize]{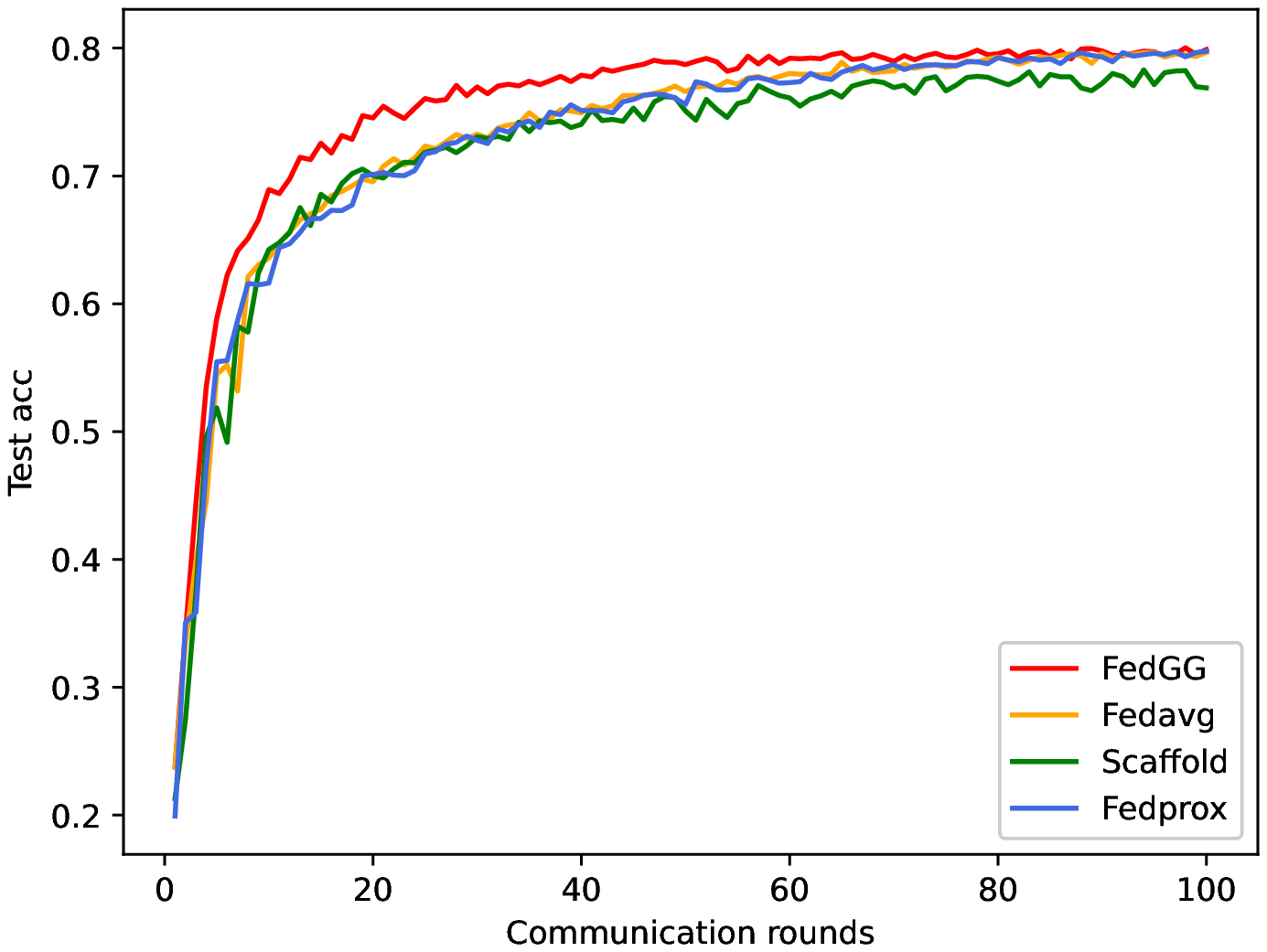}
		\end{minipage}%
	}%
	\subfigure[CIFAR-10]{
		\begin{minipage}[h]{0.3\linewidth}
			\centering
			\includegraphics[width=\hsize]{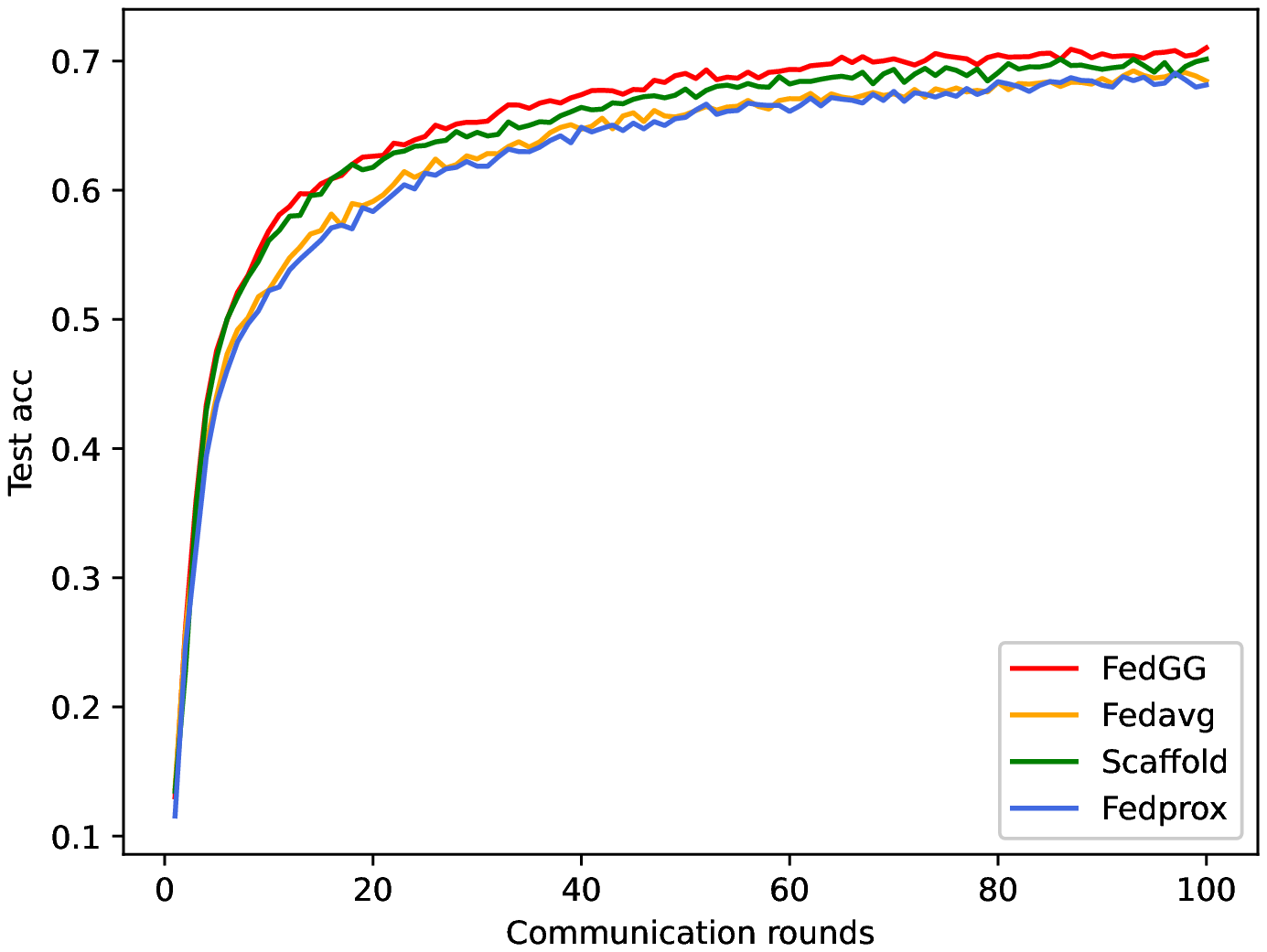}
		\end{minipage}%
	}%
	\subfigure[CIFAR-100]{
		\begin{minipage}[h]{0.3\linewidth}
			\centering
			\includegraphics[width=\hsize]{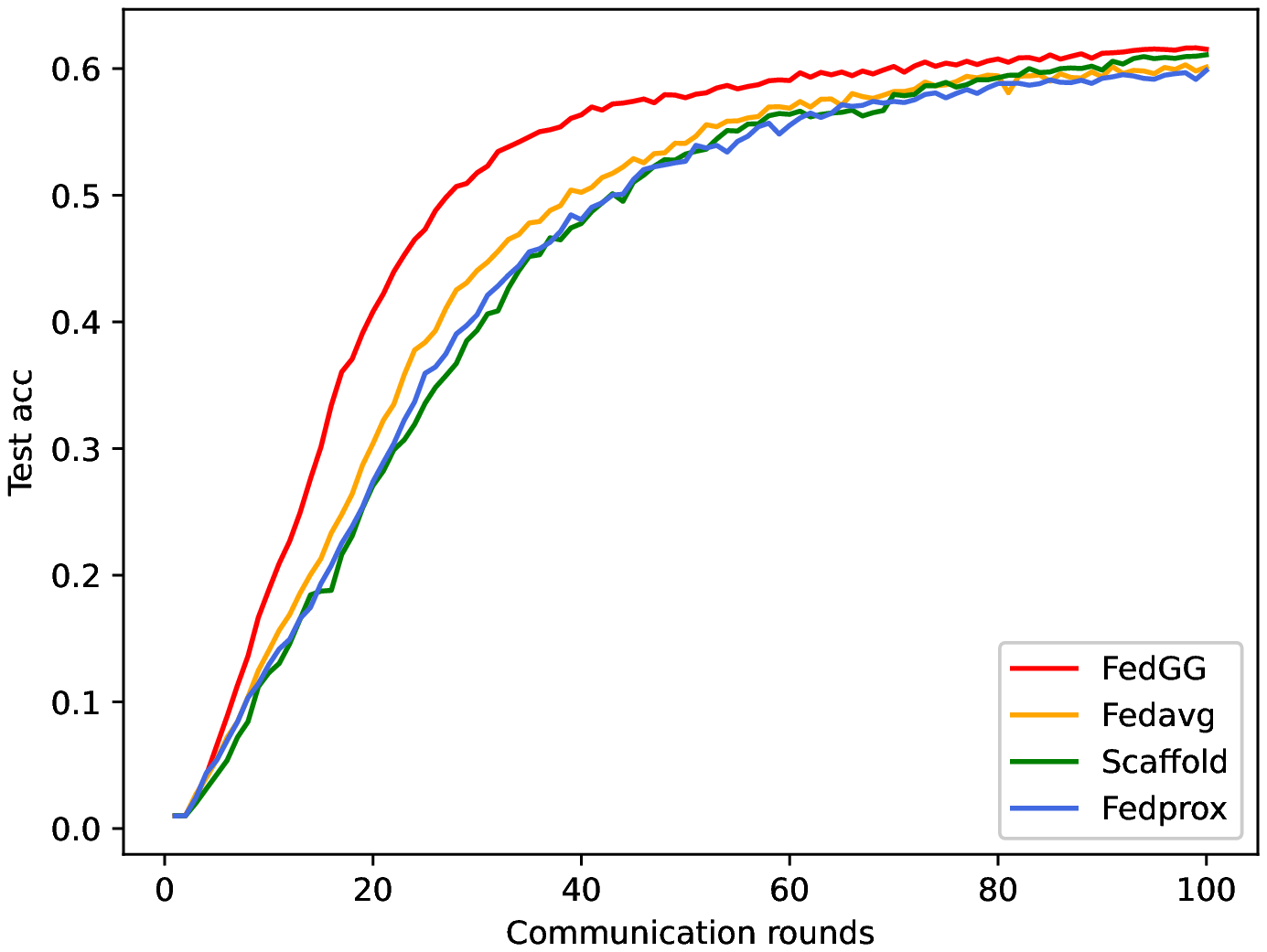}
		\end{minipage}
	}%
\\
	\subfigure[SVHN]{
	\begin{minipage}[h]{0.3\linewidth}
		\centering
		\includegraphics[width=\hsize]{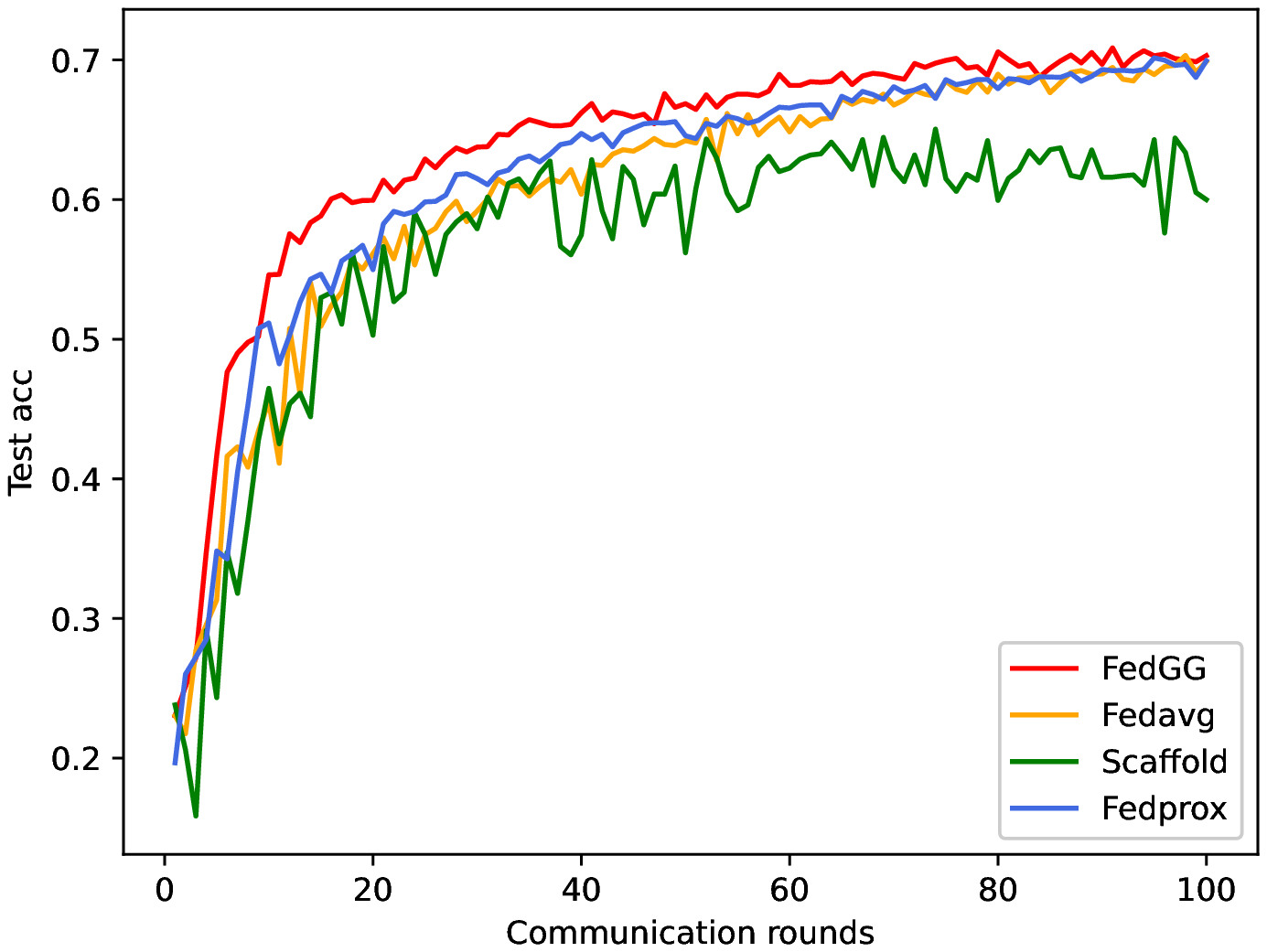}
	\end{minipage}%
}%
\subfigure[CIFAR-10]{
	\begin{minipage}[h]{0.3\linewidth}
		\centering
		\includegraphics[width=\hsize]{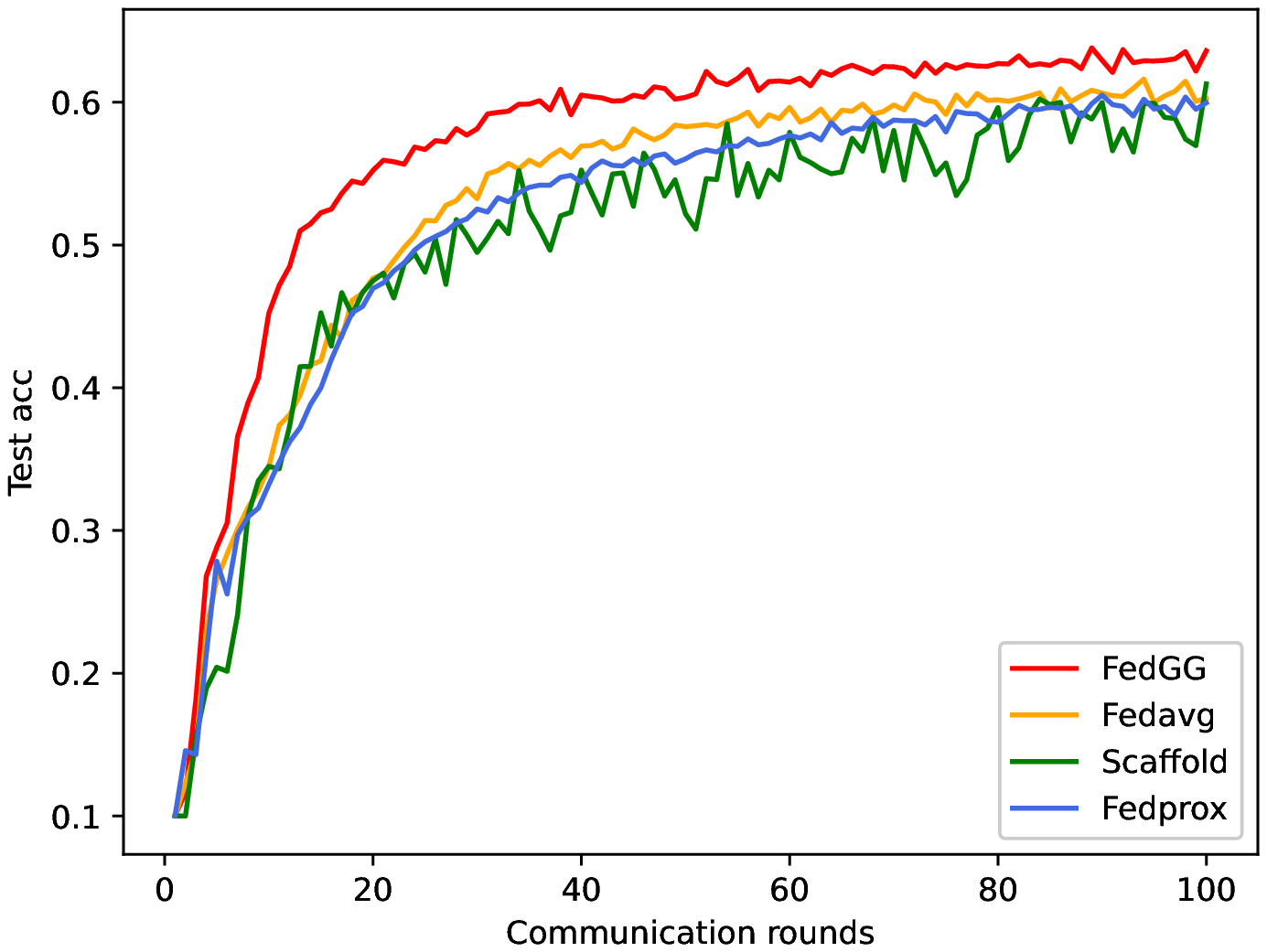}
	\end{minipage}%
}%
\subfigure[CIFAR-100]{
	\begin{minipage}[h]{0.3\linewidth}
		\centering
		\includegraphics[width=\hsize]{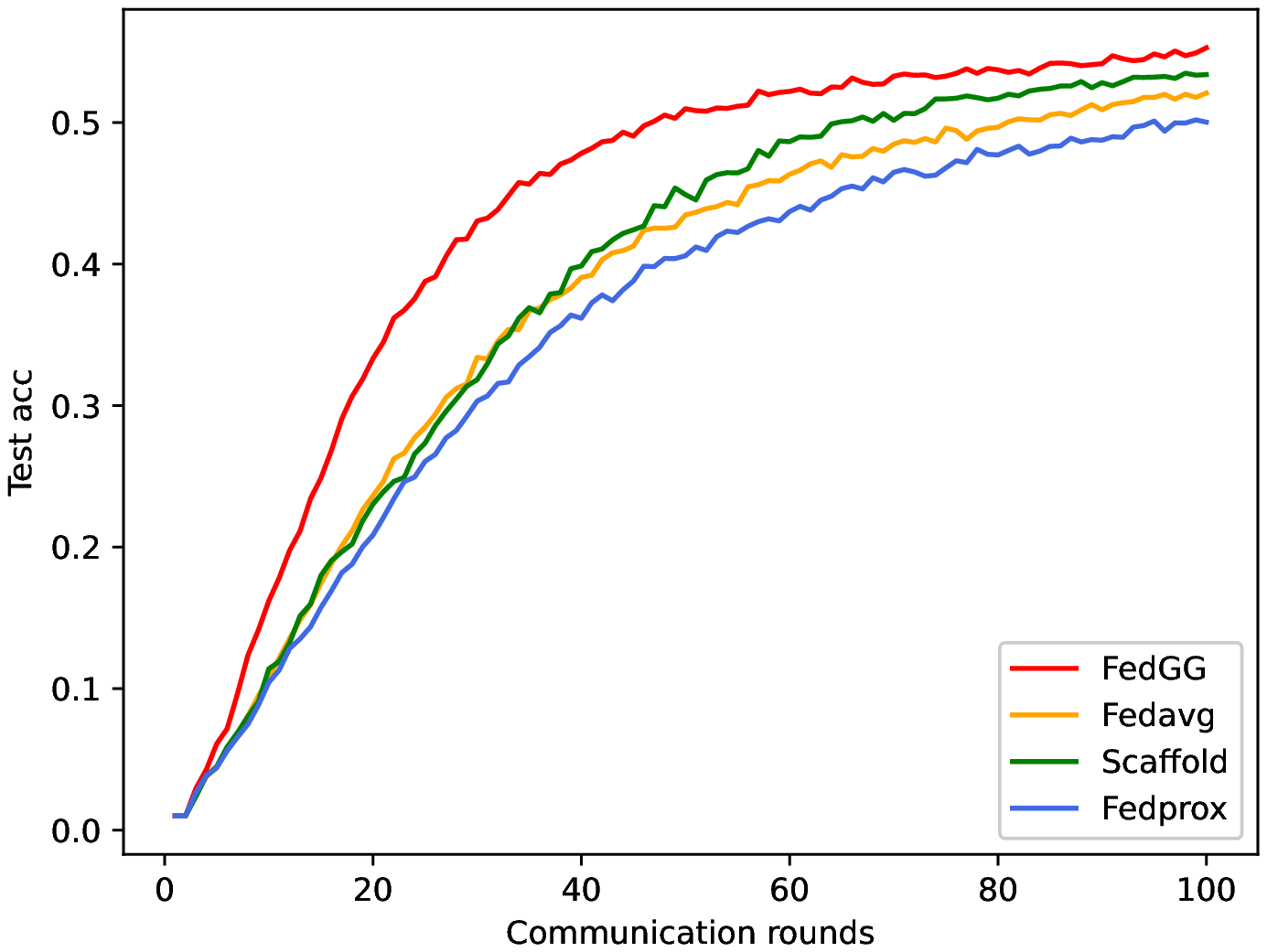}
	\end{minipage}
}%
	\centering
	\caption{Test accuracy in different number of communication rounds, for (a)-(c) $\beta=0.5$ and for (d)-(f) $\beta=0.1$.}
	\label{fig2}
\end{figure*}

The communication efficiency issue is extremely concerned in federated learning. In Fig.~\ref{fig2} we show the test accuracy in each round for $\beta=0.5$ and $\beta=0.1$. As it shown, compared to the other three algorithms, FedGG has a certain degree of improvement on the convergence accuracy and speed. Especially in the early stage of training, the accuracy performance of FedGG is significantly  better than that of the other algorithms, which greatly promotes the reduction of the communication costs. Similarly, in the case of $\beta=0.1$, due to the model-cosine loss, the advantage of FedGG becomes obvious. It is worth mentioning that the optimal hyperparameter in FedProx is usually little, so it performs very similar to FedAvg. Scaffold introduces control variables, causing the data uploaded in each communication round is twice of the other algorithms.

In order to show the high communication efficiency of FedGG, we take the algorithm test accuracy of FedAvg at the $100^{th}$ communication round as a benchmark, and list the communication rounds required for the other algorithms to achieve this accuracy in Table~\ref{tab3} and Table~\ref{tab4} with $\beta=0.5$ and $\beta=0.1$, respectively. We can observe that compared to the other algorithms, FedGG has a significant decrease in the number of communication rounds, especially on the CIFAR-10 dataset with $\beta=0.1$. The speedup reaches 2.63 times, which greatly improves the communication efficiency.
\begin{table}[htb]
	\centering
	\caption{Comparison on communication rounds for the four algorithms to achieve the accuracy of FedAvg at Round 100 in the case of $\beta=0.5$ }
	\label{tab3}
\resizebox{\linewidth}{!}{
	\begin{tabular}{c|c|c|c|c|c|c}
		\hhline
		\multirow{2}{*}{Method} &\multicolumn{2}{c}{SVHN} &\multicolumn{2}{|c}{CIFAR-10} &\multicolumn{2}{|c}{CIFAR-100} \\ \cline{2-7}
	              & rounds      & speedup    & rounds      & speedup   & rounds   & speedup\\ \hline
		FedAvg    &100          &1$\times$   &100          &1$\times$  &100       &1$\times$  \\ \hline
		Fedprox   &88           &1.13$\times$   &80          &1.25$\times$  &\diagbox{}{}       &<1$\times$  \\ \hline
		Scaffold  &\diagbox{}{} &<1$\times$   &59          &1.69$\times$  &89       &1.12$\times$  \\ \hline
		FedGG     &$\bm{65}$           &$\bm{1.53}\times$   &$\bm{47}$          &$\bm{2.12}\times$  &$\bm{70}$       &$\bm{1.42}\times$  \\ \hline
	\end{tabular}}
\end{table}
\begin{table}[htb]
	\centering
	\caption{Comparison on communication rounds for the four algorithms to achieve the accuracy of FedAvg at Round 100 in the case of $\beta=0.1$ }
	\label{tab4}
	\resizebox{\linewidth}{!}{
	\begin{tabular}{c|c|c|c|c|c|c}
		\hhline
		\multirow{2}{*}{Method} &\multicolumn{2}{c}{SVHN} &\multicolumn{2}{|c}{CIFAR-10} &\multicolumn{2}{|c}{CIFAR-100} \\ \cline{2-7}
		          & rounds      & speedup    & rounds      & speedup   & rounds   & speedup\\ \hline
		FedAvg    &100          &1$\times$   &100          &1$\times$  &100       &1$\times$  \\ \hline
		Fedprox   &95           &1.05$\times$   &90          &1.11$\times$  &\diagbox{}{}       &<1$\times$  \\ \hline
		Scaffold  &\diagbox{}{} &<1$\times$   &100          &1$\times$  &81       &1.23$\times$  \\ \hline
		FedGG     &$\bm{75}$           &$\bm{1.33}\times$   &$\bm{38}$          &$\bm{2.63}\times$  &$\bm{57}$       &$\bm{1.75}\times$  \\ \hline
	\end{tabular}}
\end{table}

\subsection{Adaptive Loss Weights}

\begin{figure}[H]
	\centering
	\subfigure[Adaptive weights with different $\mu$]{
		\begin{minipage}[H]{0.5\linewidth}
			\centering
			\includegraphics[width=\hsize]{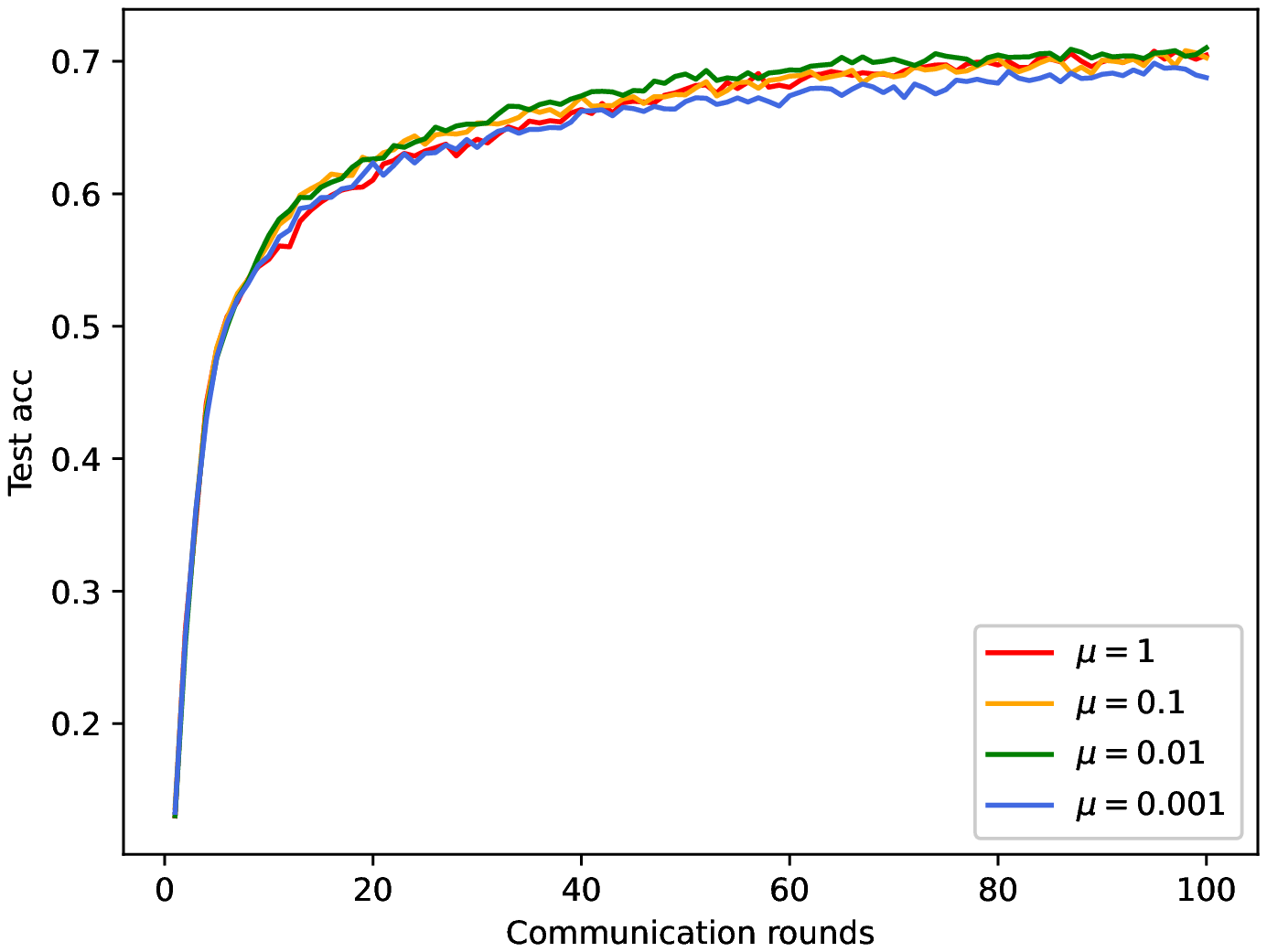}
		\end{minipage}%
	}%
	\subfigure[Different fixed weights $\lambda$]{
		\begin{minipage}[H]{0.5\linewidth}
			\centering
			\includegraphics[width=\hsize]{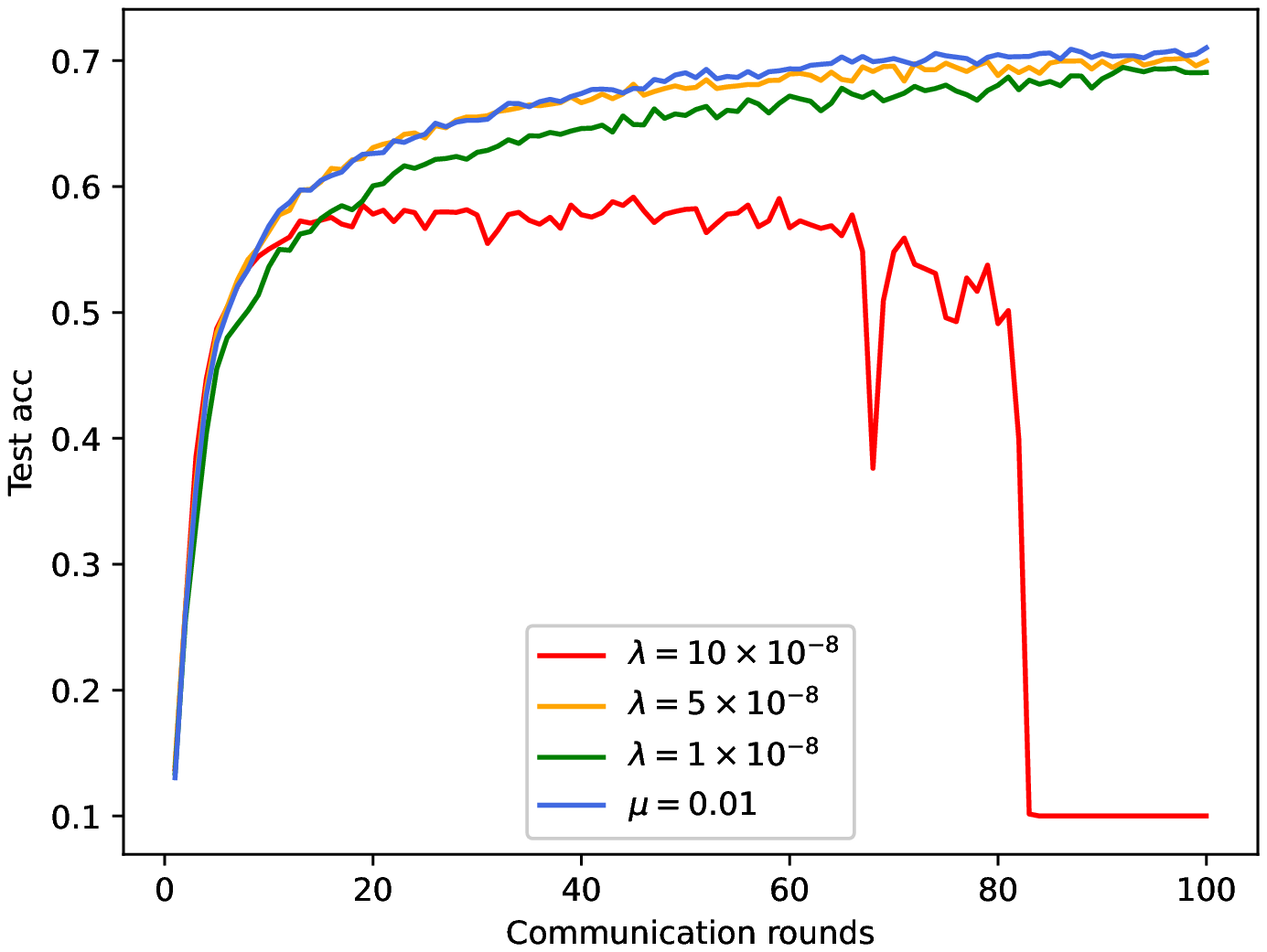}
		\end{minipage}%
	}%
	\centering
	\caption{Test accuracy for different parameters of the adaptive and fixed weights varying with the communication rounds on CIFAR-10 dataset with $\beta=0.5$.}
	\label{fig3}
\end{figure}

To investigate the effect of the adaptive loss weights on model training, we test our algorithm in CIFAR-10 dataset with $\beta=0.5$ and set the coefficient of the adaptive loss weights as $\mu=\{1,0.1,0.01,0.001\}$ and the fixed-value loss weights as $\lambda=\{10\times10^{-8},5\times10^{-8},1\times10^{-8}\}$. The variations of the test accuracy with communication rounds is shown in Fig.~\ref{fig3}. As it is shown, the best performances for the adaptive and fixed weights are $\mu=0.01$ and $\lambda=5\times10^{-8}$, respectively, and they perform very close to each other. The algorithm is stable under adaptive loss weights with different $\mu$. However, the test accuracy diverges when the fixed-value weight $\lambda=10\times10^{-8}$ which is only twice of the value for the best performance. This means that the adaptive weight has a larger adjustable range of the parameters compared to the fixed-value weights, which makes it more stable and easier to implement in practice. Since in the actual situation that the server generally has no test data, the global model can only be evaluated when the local training is over. At this time, a larger adjustable range of the parameter can be better tolerance for failure. Furthermore, although little values of both fixed-value weights and adaptive weights calculated from Eq.~(\ref{eq10}) indicate that the model-cosine loss has a little impact on the total local loss, the large order of magnitude for the gradient of the model-cosine loss makes it reasonably affect the local update.

\section{Conclusion}
In this paper, we have proposed global-update-guided federated learning (FedGG). By introducing model-cosine loss, the local model can fit the local data while update closely along the direction of the global model. This reduces the divergence between the local and global models and improves the performance of the model on Non-IID data. Besides, by introducing the adaptive loss weights, the adjustable range of model parameters is increased, making it easier and more stable to implement in practice. Our experiments show that, compared with the other state-of-the-art algorithms, FedGG's accuracy increases by $0.5\%$, $2.2\%$ and $2.9\%$ on Non-IID SVHN, CIFAR-10, and CIFAR-100 datasets. And the speedup in communication efficiency is up to 2.63 times while not increasing the amount of communication per round.


\begin{thebibliography}{0}
\bibitem{Yang}
Q.~Yang, Y.~Liu, T.~Chen, et al, Federated Machine Learning: Concept and Applications, \emph{ACM Transactions on Intelligent Systems and Technology (TIST)}, 10(2): 1--19, 2019.
\bibitem{Kumar}
M.~Zhang, Y.~Wang, T.~Luo, Federated Learning for Arrhythmia Detection of Non-IID ECG, \emph{IEEE 6th International Conference on Computer and Communications (ICCC)}, 2020: 1176--1180.
\bibitem{Guan}
Y.~Guan, D.~Li, S.~Xue, et al, Feature-Fusion-Kernel-Based Gaussian Process Model for Probabilistic Long-Term Load Forecasting, \emph{Neurocomputing}, 426(2): 174--184, 2021.
\bibitem{Jia}
S.~Jia, Z.~Gan, Y.~Xi, et al, A Deep Reinforcement Learning Bidding Algorithm on Electricity Market, \emph{Journal of Thermal Science}, 29(5): 1125--1134, 2020.
\bibitem{GuanQ}
Q.~Guan, W.~Li, S.~Xue, et al, High-Resolution Representation Object Pose Estimation from Monocular Images, \emph{Chinese Automation Congress}, 2021: 980--984.
\bibitem{GuanQ1}
Q.~Guan, Z.~Sheng and S.~Xue, HRPose: Real-Time High-Resolution 6D Pose Estimation Network Using Knowledge Distillation, \emph{Chinese Journal of Electronics}, accepted, 2022.
\bibitem{Sheng}
Z.~Sheng, S.~Xue, Y.~Xu, et al, Real-Time Queue Length Estimation With Trajectory Reconstruction Using Surveillance Data, \emph{2020 16th International Conference on Control, Automation, Robotics and Vision (ICARCV)}, 2020: 124--129.
\bibitem{Sheng1}
Z.~Sheng, Y.~Xu, S.~Xue, et al, Graph-Based Spatial-Temporal Convolutional Network for Vehicle Trajectory Prediction in Autonomous Driving, \emph{IEEE Transactions on Intelligent Transportation Systems}, online published, 2022. DOI: 10.1109/TITS.2022.3155749
\bibitem{Sheng2}
Z.~Sheng, L.~Liu, S.~Xue, et al, A Cooperation-Aware Lane Change Method for Autonomous Vehicles, \emph{ArXiv Preprint}, arXiv: 2201.10746, 2022.
\bibitem{AbdulRahman}
S.~AbdulRahman, H.~Tout, H.~Ould-Slimane, et al, A Survey on Federated Learning: The Journey from Centralized to Distributed On-Site Learning and Beyond. \emph{IEEE Internet of Things Journal}, 8(7): 5476-5497, 2020.
\bibitem{Voigt}
P.~Voigt and A.Bussche, The EU General Data Protection Regulation (GDPR), \emph{A Practical Guide, 1st Ed.}, Cham: Springer International Publishing, 2017.

\bibitem{McMahan}
B.~McMahan, E.~Moore, D.~Ramage, et al, Communication-Efficient Learning of Deep Networks from Decentralized Data, \emph{Artificial Intelligence and Statistics}, 2017: 1273-1282.

\bibitem{Li}
T.~Li, K.~Sahu, A.~Talwalkar, et al, Federated Learning: Challenges, Methods, and Future Directions, \emph{IEEE Signal Processing Magazine}, 37(3): 50-60, 2020.
\bibitem{Lix}
X.~Li, K.~Huang, W.~Yang, et al, On the Convergence of FedAvg on Non-IID Data, \emph{International Conference on Learning Representations (ICLR)}, 2019: 1-26.
\bibitem{Li1}
T.~Li, A.~Sahu, M.~Zaheer, et al, Federated Optimization in Heterogeneous Networks, \emph{Proceedings of Machine Learning and Systems}, 2020: 429-450.
\bibitem{Shoham}
N.~Shoham, T.~Avidor, A.~Keren, et al, Overcoming Forgetting in Federated Learning on Non-IID Data. \emph{NeurIPS Workshop}, 2019: 1-6.
\bibitem{Hou}
S.~Hou, X.~Pan, C.~Loy, et al, Lifelong Learning via Progressive Distillation and Retrospection, \emph{In Proceedings of the European Conference on Computer Vision (ECCV)}, 2018: 437-452.
\bibitem{Yao}
X.~Yao , L.~Sun, Continual Local Training for Better Initialization of Federated Models, \emph{2020 IEEE International Conference on Image Processing (ICIP)}, 2020: 1736-1740.
\bibitem{Karimireddy}
S.~Karimireddy, S.~Kale, M.~Mohri, et al, SCAFFOLD: Stochastic Controlled Averaging for Federated Learning. \emph{In International Conference on Machine Learning}, 2020: 5132-5143.
\bibitem{Li3}
Q.~Li, B.~He, D.~Song, Model-Contrastive Federated Learning, \emph{Proceedings of the IEEE/CVF Conference on Computer Vision and Pattern Recognition (CVPR)}, 2021: 10713-10722.
\bibitem{LiB}
B.~Li, L.~Han, Distance Weighted Cosine Similarity Measure for Text Classification, \emph{International Conference on Intelligent Data Engineering and Automated Learning}, 2013: 611-618.
\bibitem{Geyer}
R.~Geyer, T.~Klein, M.~Nabi, Differentially Private Federated Learning: A Client Level Perspective, \emph{NeurIPS Workshop}, 2017: 1-7.


\end{thebibliography}
\end{document}